\documentclass[english]{llncs}
\usepackage{babel}
\usepackage{array}
\usepackage{multirow}
\usepackage{amsmath}
\usepackage{graphicx}
\usepackage{comment}
\usepackage{longtable}
\usepackage{tabularx}
\usepackage{adjustbox}
\usepackage{hhline}
\usepackage{xcolor}
\usepackage[utf8]{inputenc}
\usepackage{multirow,booktabs}
\usepackage{amsmath}
\usepackage{comment}
\usepackage{subfig}

\usepackage[]{pdfpages}

\usepackage{hyperref}
\usepackage{url}
\usepackage{tabularx}
\usepackage{graphicx}
\usepackage{tabularx}
\usepackage{adjustbox}
\usepackage{hhline}
\usepackage{graphicx}
\usepackage{color,soul}
\usepackage[toc,page]{appendix}

\makeatletter

\providecommand{\tabularnewline}{\\}

\usepackage{microtype}
\usepackage{lmodern}

\makeatother

\begin{document}

\title{DeepProcess: Supporting business process execution using a MANN-based recommender system}

\author{Asjad Khan$^{1,}$, Hung Le$^{2,*}$, Kien Do$^{2}$, Truyen Tran$^{2}$, Aditya Ghose$^{1}$, Hoa Dam$^{1}$, and Renuka Sindhgatta$^{3}$ }

\institute{
$^{1}$University of Wollongong, Australia\\
\texttt{maak458@uowmail.edu.au; \{aditya,hoa\}@uow.edu.au}~\\
	$^{2}$Deakin University, Geelong, Australia\\
\texttt{\{lethai,dkdo,truyen.tran\}@deakin.edu.au}\\
$^{3}$IBM Research India, India\texttt{}~\\
\texttt{renuka.sr@in.ibm.com}}

\maketitle

\begin{abstract}
Process-aware Recommender systems can provide critical decision support functionality to aid business process execution by recommending what actions to take next. Based on recent advances in the field of deep learning, we present a novel memory-augmented neural network (MANN) based approach for constructing a process-aware recommender system. We propose a novel network architecture, namely Write–Protected Dual Controller Memory–Augmented Neural Network(DCw-MANN), for building prescriptive models.  To evaluate the feasibility and usefulness of our approach, we consider three real-world datasets and show that our approach leads to better performance on several baselines for the task of suffix recommendation and next task prediction.

\end{abstract}

\global\long\def\xb{\boldsymbol{x}}
\global\long\def\yb{\boldsymbol{y}}
\global\long\def\eb{\boldsymbol{e}}
\global\long\def\zb{\boldsymbol{z}}
\global\long\def\hb{\boldsymbol{h}}
\global\long\def\ab{\boldsymbol{a}}
\global\long\def\bb{\boldsymbol{b}}
\global\long\def\cb{\boldsymbol{c}}
\global\long\def\sigmab{\boldsymbol{\sigma}}
\global\long\def\gammab{\boldsymbol{\gamma}}
\global\long\def\alphab{\boldsymbol{\alpha}}
\global\long\def\rb{\boldsymbol{r}}
\global\long\def\gb{\boldsymbol{g}}
\global\long\def\Deltab{\boldsymbol{\Delta}}
\global\long\def\hb{\boldsymbol{h}}
\global\long\def\prefix{\text{prefix}}
\global\long\def\suffix{\text{suffix}}
\global\long\def\DLev{D_{\text{Lev}}}
\global\long\def\RNN{\text{RNN}}
\global\long\def\Loss{\mathcal{L}}
\global\long\def\Neigh{\mathcal{N}}
\global\long\def\error{\text{error}}
\global\long\def\fb{\boldsymbol{f}}
\global\long\def\ib{\boldsymbol{i}}
\global\long\def\ob{\boldsymbol{o}}
\global\long\def\kb{\boldsymbol{k}}
\global\long\def\wb{\boldsymbol{w}}
\global\long\def\vb{\boldsymbol{v}}

\section{Introduction}

Business process management assists organizations in planning and executing activities that collectively deliver business value, usually in the form of a product or a service. Flexible execution of business process instances entails multiple critical decisions, involving various actors and objects, which can have a major impact process performance and achieving desired process outcomes \cite{teinemaa2019outcome}. These decisions therefore require careful attention, as sub-optimal decisions during process execution, can lead to cost overruns, missed deadlines and the risk of failure \cite{ghattas2014improving}. While the problem of predicting the behaviour of a given process instance has been studied extensively, using these predictions to support operational decision-making of the kinds outlined above
remains a challenge \cite{marquez2017predictive}\cite{dees2019if}. \textit{Process-Aware Recommender Systems} have been proposed to assist knowledge workers in operational decision-making, for instance, by recommending actions leading to process end, managing resource allocation policies and so on \cite{beheshti2020towards} \cite{schonenberg2008supporting}. In this work, we present a novel \emph{Process-Aware Recommender System} for supporting organizations and process owners in operational decision-making (related to control-flow).  


Recent advances in neural network architectures and learning algorithms have led to the popularization of \emph{Deep Learning} methods which
are particularly good at automated feature discovery and learning robust representations from large quantities of raw data, thus significantly reducing the need to hand-craft features which is typically required when using traditional machine learning techniques \cite{lecun2015deep}. Deep Learning based techniques such as
Long Short-Term Memory (LSTM) and Gated Recurrent Units (GRUs) have
generated considerable interest recently for tackling various Process Analytics tasks (e.g predictive monitoring). However, LSTMs and GRU methods lack the capacity to solve complex, structured tasks that, for example, require reasoning and  planning\cite{evermann2017predicting} \cite{graves2016hybrid}. To tackle such complex tasks, two promising approaches based on neural networks have been proposed: Memory Networks and Neural Turing Machines, both being instantiations of \emph{Memory-Augmented Neural Networks} (MANN)\cite{graves2014neural}. In this paper, we investigate the applicability of MANNs for building a \emph{Process-Aware Recommender System} that can provide process execution decision support of the kind discussed above. \\

\noindent
\textbf{Contributions:} We propose a novel neural network architecture, namely\emph{Write–Protected Dual Controller Memory–Augmented Neural Network(DCw-MANN)}, for building a Process-Aware Recommender System, where we introduce several modifications to the existing Differential Neural Computer(DNC) architecture: (i) \emph{separating the encoding phase and decoding phase, resulting in dual controllers, one for each phase}; (ii) \emph{implementing a write-protected policy for memory during the decoding phase}. We evaluate the effectiveness of our approach on three world datasets for the task of \emph{generating suffix recommendations} that lead to optimal outcomes. 

The paper is organized as follows: 
In Section~\ref{sec:Preliminaries}, we provide the necessary background on Process Analytics and Deep Learning techniques upon which our proposed method is built. In Section~\ref{sec:Methods}, we explain the technical workings of our Process-Aware Recommender System, designed to tackle a number of prescriptive process analytics tasks. Implementation details
and experimental results are reported in Section~\ref{sec:Evaluation}. Finally, Section~\ref{sec:Related-Work}
discusses related work, followed by Section~\ref{sec:Conclusions} which concludes the paper and outlines future work.

\section{Preliminaries\label{sec:Preliminaries}}
We first briefly present the existing work upon which our method is
built, including event log presentation, recurrent neural networks, and Long Short-Term Memory (LSTM).

\subsection{Process Analytics}

Process analytics involves a sophisticated layer of data analytics built over the traditional notion of process mining \cite{van2011process}. Compared to Process mining, Process analytics addresses the more general problem of leveraging data generated by, or associated with, process execution to obtain actionable insights about business processes. Process analytics leverages a range of data, including, but not limited to process logs, event logs \cite{santiputri2017mining}, provisioning logs, decision logs and process context \cite{sindhgatta2016context} and answers queries that have a number of real world applications particularly related to prescriptive analytics such as resource optimisation and instance prioritisation. In this paper we focus on event logs and assume that when a business process instance is executed, its execution
trace is recorded as an event log.  An event log is a sequence of events, naturally ordered by the associating timestamps. 

In predictive analytics, we study techniques that allow us to predict how the future of a given process instance will unfold and the likely occurrence of future process events \cite{evermann2017predicting}. It can be considered as computing (a)
a set of functions and (b) a set of computer programs that carry out
computation, over a (partially executed) process instance. An example
of case (a) is computing remaining time of a process instance, which
is the \emph{sequence-to-vector} setting. An example of case (b) is
a continuation of a partially executed process, which is the \emph{sequence-to-sequence}
setting. 

Prescriptive business process monitoring techniques and Process Aware Recommender systems are for providing decision-support to process users. Applications of such system include, offering recommendations about:   \emph{(i)} next activities to execute,  \emph{(ii)} resource allocation support,  \emph{(iii)} Cost and time optimization and  \emph{(iv)} risk-mitigation by raising alarms or recommending actions to prevent undesired outcomes\cite{weinzierl2020prescriptive}\cite{eili2021systematic}. 

\subsection{Sequence Modeling with Deep Learning}

Recurrent Neural Nets(RNNs), especially the Long Short-Term Memory (LSTM) have brought about breakthroughs in solving complex sequence modelling tasks in various domains such as video understanding, speech recognition and natural language processing \cite{schmidhuber2015deep,lecun2015deep}. Similarly, it has been shown that LSTM can consistently outperform classical techniques for a number of process analytics tasks such as predicting the next activity, time to the next activity etc.\cite{tax2017predictive,navarin2017lstm}.


Recurrent neural network (RNN) is a model of dynamic processes, and
to some degree, a model of computer programs. At each time step $t$,
a RNN reads an \emph{input vector} $\xb_{t}$ into a \emph{hidden
state vector} $\hb_{t}$ and predicts an \emph{output vector} $\yb_{t}$.
The state dynamic can be abstracted as a recurrent relation: $\hb_{t}=\text{RNN}\left(\hb_{t-1},\xb_{t}\right)$.
The vanilla RNN is parameterized as follows:

\begin{align*}
\hb_{t} & =\sigma\left(W_{h}\hb_{t-1}+V\xb_{t}+\bb_{h}\right)\\
\yb_{t} & =W_{y}\hb_{t}+\bb_{y}
\end{align*}
where $\left(W_{h},W_{y},V,\bb_{h},\bb_{y}\right)$ are learnable
parameters, and $\sigma$ is a point-wise nonlinear function.

Although theoretically powerful, vanilla RNNs cannot learn from long-sequences
due to a problem known as vanishing or exploding gradients. A powerful
solution is Long Short-Term Memory (LSTM) \cite{hochreiter1997long}.
LSTM introduces one more vector called ``memory'' $\cb_{t}$,
which, together with the state $\hb_{t}$, specify the dynamic as:
$\left(\hb_{t},\cb_{t}\right)=\text{LSTM}\left(\hb_{t-1},\cb_{t-1},\xb_{t}\right)$.
In most implementations, this is decomposed further as:
\begin{align*}
\cb_{t} & =\fb_{t}\ast\cb_{t-1}+i_{t}\ast\tilde{\cb}_{t}\\
\hb_{t} & =\ob_{t}\ast\text{tanh}\left(\cb_{t}\right)
\end{align*}
where $\tilde{\cb}_{t}$ is a candidate memory computed from the input,
$\fb_{t},\ib_{t},\ob_{t}\in(\boldsymbol{0},\boldsymbol{1})$ are gates,
and $\ast$ denotes point-wise multiplication. $\fb_{t}$ determines
how much the previous memory is maintained; $\ib_{t}$ controls how
much new information is stored into memory, and $\ob_{t}$ controls
how much memory is read out. The candidate memory and the gates are
typically parameterized as:
\begin{align*}
\tilde{\cb}_{t} & =\text{tanh}\left(W_{c}\hb_{t-1}+V_{c}\xb_{t}+\bb_{c}\right)\\
\left[\begin{array}{c}
\fb_{t}\\
\ib_{t}\\
\ob_{t}
\end{array}\right] & =\text{sigm}\left(\left[\begin{array}{c}
W_{f}\\
W_{i}\\
W_{o}
\end{array}\right]\hb_{t-1}+\left[\begin{array}{c}
V_{f}\\
V_{i}\\
V_{o}
\end{array}\right]\xb_{t}+\left[\begin{array}{c}
\bb_{f}\\
\bb_{i}\\
\bb_{o}
\end{array}\right]\right)
\end{align*}
where $\left(W_{c,f,i,o},V_{c,f,i,o},\bb_{c,f,i,o}\right)$ are learnable
parameters.

\section{Approach\label{sec:Methods}}

While LSTMs can theoretically deal with long event sequences, the long-term dependencies between distant events in a process get diffused into the memory vector.  LSTM partly solves the gradient issue associated with the vanilla RNN but it may not be very effective on complex process executions that contain multiple computational steps
and long-range dependencies. Keeping this in mind, we explore the application of an expressive sequential process model, that would allow storing and retrieval of intermediate process states in a long-term memory. This is akin to the capability of a trainable Turing machine. Closest to a Turing machine is an instantiation of MANN, known as Differential Neural Computer (DCN) \cite{graves2016hybrid}.  MANNs can be considered as a recurrent net augmented with an external memory module \cite{graves2016hybrid,sukhbaatar2015end}.
Because of this memory module MANNs have certain advantages over traditional LSTMs when tackling highly complex sequence modeling problems such as question answering \cite{sukhbaatar2015end} and algorithmic tasks \cite{graves2016hybrid}. The memory $\cb_{t}$ compresses the entire history
into a single vector, and thus the process structure is somewhat lost.
For example, if two distant events are highly dependent, there are
no easy ways to enforce this relationship through the forgetting gates.
Another critical issue is that if a process involves multiple intermediate results
for latter use, there are no mechanism to store these results into
the flat memory vector $\cb_{t}$. These drawbacks demand an external
memory to store temporary computational results, akin to the role
of RAM in modern computers. The key idea behind these architectures is that all memory operations,
including addressing, reading and writing are differentiable. This
enables end-to-end gradient-based training. MANNs have found many
applications, e.g., question answering \cite{kumar2016ask,sukhbaatar2015end}
and simple algorithmic tasks \cite{graves2014neural}. Overall, Encoder-decoder architectures like \emph{memory-augmented
neural nets} are geared to solve sequence to-sequence problems and are naturally a good fit for tackling the problem of \emph{optimal path recommendation}.  

We adapted the most advanced variant of MANNs to
date, the Differential Neural Computer (DNC) \cite{graves2016hybrid}. In most popular implementations, DNC can be considered as a LSTM augmented with an external memory module $M$. The LSTM plays the role of a controller, which is akin to a CPU, where
the memory $\cb_{t}$ is akin to registers in the CPU. At each time
step, the controller (i) reads an input, (ii) updates its own internal
memory and states, (iii) writes the new information data into the
external memory, and (iv) finally reads the updated memory to produce
an output. In a typically implementation, the external memory is a matrix of $N$ slots, each slot is a vector. To interface with the external memory, the controller computes keys $\kb_{t}$ for locating slots for reading and writing.
The memory slot is found using cosine similarity
between the key and the slot content. This mechanism of locating memory slot is known as \emph{content-based addressing}. In addition, DNC also supports \emph{dynamic memory allocation}
and \emph{temporal memory linkage} mechanisms for computing one final write-weight and several read-weights.

The read-weights are then used to produce a read content
from the memory. Multiple reads are then combined with the controller state to produce an output vector $\ob_{t}$. For readability, we omit the mathematical
details here. Readers are referred to the original paper \cite{graves2016hybrid}.

We now describe how the DNC can be adapted for prescriptive process analytic tasks, starting from event coding into the model and decoding from it, to specific modifications of the DNC to make it suitable for solving a variety of prescriptive tasks in business processes.

\subsection{Events/resources coding and decoding \label{subsec:Events/resources-coding-and}}

Discrete events/resources in event log can be coded into MANN in several
ways. If the number of unique events/resources is large, embedding
into a low-dimensional space is typically employed, that is $a\rightarrow\xb_{a}$.
Otherwise, a simple one-hot coding will suffice, that is, $a\rightarrow\left[0,0,...1,...0\right]$.
Continuous resources such as time can be normalized as input variables.
Alternatively, these continuous variables can be discretized into
symbols that represent intervals. This could enable true \emph{end-to-end}
learning. However, we can also employ a certain degree of feature
engineering to enhance the input signals as in \cite{tax2017predictive},
which has been shown to be highly effective.

For discrete symbol prediction at time $t$, we can use a softmax:
\begin{equation}
P_{t}\left(a\mid\text{history}\right)=\frac{\exp\left(\wb_{a}\cdot\ob_{t}\right)}{\sum_{a'}\exp\left(\wb_{a'}\cdot\ob_{t}\right)}\label{eq:event-decode}
\end{equation}
where $\ob_{t}$ is the output vector generated by the controller,
and $\wb_{a}$ is a trainable parameter vector. The discrete output is simply: $a^{*}=\arg\max_{a}P_{t}\left(a\mid\text{history}\right)$.
Continuous prediction is through a function $y_{t}=f\left(\ob_{t}\right)$,
which can be itself a feedforward neural net.

Application of these decoding settings in  Eq.~(\ref{eq:event-decode}) allows us to solve a variety of predictive and prescriptive tasks like  next-activity recommendation, suffix recommendation and so on. Likewise time-to-event
estimation is simply continuous prediction.

\subsection{Sequence prediction with dual-controllers and write-protected policy}

We assume that at the decision point, we are given a partially
executed process instance, and we want to prescribe actions for the continuation of a process instance based and optimize for KPIs based on remaining time, or the set of resources needed for completing the instance. Under the MANN formulation, many of those prescriptive tasks can be cast into \emph{sequence prediction}, that is, we generate a sequence of discrete symbols. For example, process continuation is a natural case, where each symbol is an event.

In case of resources prediction, even though there may or may not
natural ordering among resources, we can still produce a sequence.
Due to the availability of the external memory which stores all the previous knowledge, the strict ordering in the output sequence is not of a major issue, because at any point in the prediction time, the controller can just make use of the external memory (which can be order-free since it is read-only), and relies less on its own internal memory (which is order-dependent). Note that this property is not possible in LSTM, which is sequential by design.

In the DNC setting, this task can be decomposed into dual phases:
the \emph{encoding phase}, in which the prefix is read into the memory,
and the \emph{decoding phase}, in which the suffix is sequentially
generated. Second, in standard DNC operations, the memory is constantly modified at each time step. In the dual-phase setting, there is no need to update the memory since there are no real inputs. Thus we suggest a simple modification, that is, the memory is read-only during the decoding phase. And finally, since the two phases serve different purposes, it might be useful to separate the \emph{encoding controller}
from the \emph{decoding controller}. That is, the encoding controller is specialized in keeping the best description of the process thus far, and the decoding controller is optimized to producing the best suffix, given the information pre-computed by the encoding controller.
We call this DNC variant \emph{DCw-MANN}, which stands for Write\textendash Protected
Dual Controller Memory\textendash Augmented Neural Network. The proposed system learns a highly compact low-dimensional process representation and captures all variations implicit in the given process execution log to enable \emph{near real-time decision support} for tackling multiple prescriptive monitoring tasks.

\begin{figure*}
	\begin{centering}
		\includegraphics[width=0.5\linewidth]{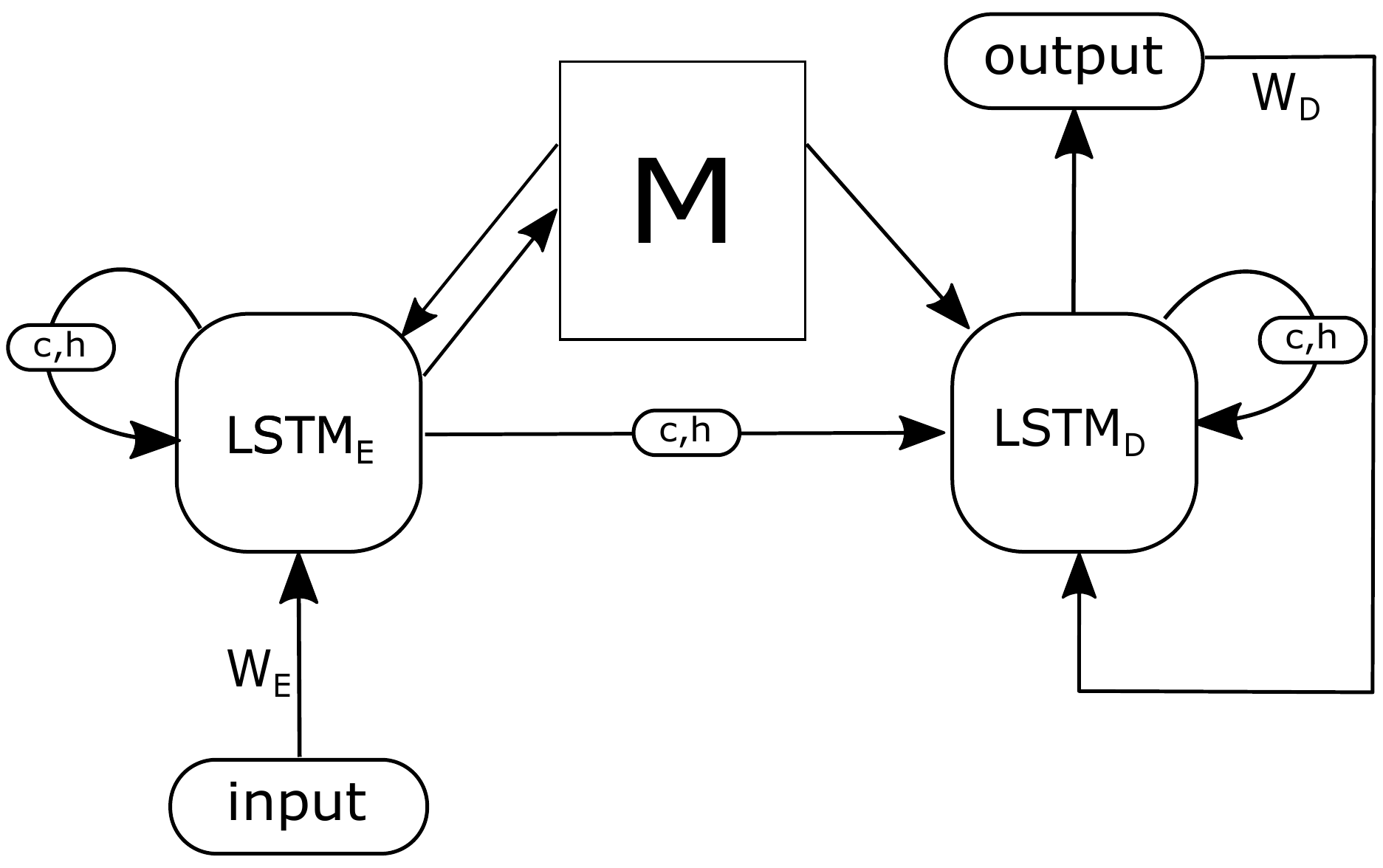}
		\par\end{centering}
	\centering{}\caption{Write-Protected Dual Controller Memory Augmented Neural Network \label{fig:Dual-Controller-Write} }
\end{figure*}

\textbf{Model operations over time:} The operations of the modified DNC is illustrated in Figure \ref{fig:Dual-Controller-Write}.
There are two controllers, the encoder $\text{LSTM}_{enc}$ for the
encoding phase and the decoder $\text{LSTM}_{dec}$ for the decoding
phase. Both share the same external memory $M$. Each controller maintains
their own internal memory $\cb$ and state $\hb$. In the encoding
phase, the prefix is fed into the encoder one event at a time. The
external memory is updated a long the way. In decoding phase, the
state of the encoder and the memory are passed into the decoder. The
long-range dependencies between the input prefix and the output suffix
are maintained through the memory look-up operations.

During the sequence decoding phase, the next symbol at time $t$ is
predicted using the information from the memory and previously generated
symbols:
\[
P_{t}\left(a\mid a_{t_{pre}+1},...,a_{t-1},\text{history}\right)
\]
as in Eq.~(\ref{eq:event-decode}), where the output $\ob_{t}$ is
generated by the decoder $\text{LSTM}_{dec}$.

\subsection{Generating Suffix Recommendations for decision-support}

Next, our we goal is to learn the task of generating optimal suffix recommendations from partially executed process instances. Process-Aware Recommender Systems, support process users in operational decision-making by continuously monitoring process executions and providing automated recommendations which maximize the likelihood of achieving desired process outcomes.   
Machine learning based approaches are commonly used to construct data-driven recommender systems where the system attempts to predict the user's interests and recommends items based on those interests. Many of the standard industry recommender systems build a machine learning model by leveraging the user's past behaviour (which is routinely logged) as well as similar actions taken by other users. This model is then used to predict items (or ratings for items) that the users may have an interest in. In Process Analytics, the operative notion of {\em recommendation}, can be realised by using a machine learning based system, capable of learning from successful (or well-performing) process instances.  Weber et. al \cite{schonenberg2008supporting} have explained outcome oriented recommendations based on predictions, as follows: ``Recommendations can be considered as predictions about a case, conditioned on the next step that has not been performed yet. In order to recommend to a user what should be the next step in a process, the recommendation service needs to know what the user’s target (goal) is, e.g. should the user perform its tasks as soon as possible, or should s/he optimize its outcome in terms of business value''. 



We have designed our Process-Aware Recommender System to consider the process outcomes by \emph{i)} implementing task conditioning at an architectural level (i.e using task specific encoders/decoders) \emph{ii)}  leveraging past execution data labelled with outcomes (based on performance indicators or non-functional attributes defined for the task). Such labelled data, contains rich knowledge  
capturing cumulative best practices from the perspective of multiple process users. Our system underpins operational decision support in a manner where good performing instances are leveraged to train a model that can correlate actions with the likelihood of their effectiveness. Here, labels help in differentiating between process instances that performed well based on a pre-defined performance criterion (e.g through-put time) versus those that performed poorly.





The training examples allow our model to learn the relevant representations from raw data.  We trained our proposed machinery in a manner similar to the task of training unsupervised language models, where sequence prediction models are trained with a simple objective: predict the next word, given all of the previous words within some text \cite{radford2019language}.  Following this approach allows us to learn the prescriptive tasks without the need for explicit supervision. Furthermore, this approach allows us to build a general-purpose model that assumes no domain-specific knowledge of the process, other than the symbolic representation of events (or resources). We finally note that, once the model is trained, it doesn't simply match or repeat the same sequence(recommendations) from training logs rather testing it on an unseen test set(as done in our experiments) shows that our model has learned the task of \emph{recommending an optimal suffix} given a partially executed instance. Good performance on test set also shows that capability of our model to generalise such that it can perform well on a wide variety of future unseen process instances(that were missing from the training data).

\section{Evaluation \label{sec:Evaluation}}

In the following sections, we explain the experimental setup, we then describe the datasets and pre-processing strategy used for evaluating our proposed approach(Section 4.2 and 4.3). We motivate the choice of metrics and describe the baseline
methods. We finally present an explanation of model implementation (Section 4.4) along with experimental results. 

\subsection{Datasets \label{subsec:Datasets}}

We consider three datasets to to evaluate our suffix recommendation engine, whose description is as follows:

\begin{itemize}

\item \textbf{Moodle Dataset}: This dataset has been created from Moodle's(e-learning platform) issue tracking system. The issue tracking system collects bug reports and allows developers to track the bug resolution process as an issue goes through various development stages. The log contains 10,219 complete processes in total with the number of events in each process ranging from 4
to 23. The preprocessing procedure results in about 32K training prefix/suffix
sequences and 8K prefix/suffix sequences. The number of event codes
in Moodle dataset is 23. 

\item \textbf{Financial Log}: This log is based on BPI2012 challenge dataset but was preprocessed(see description below) based on a time-based performance metric. After pre-processing we are only left with good performing instances which can be fed to the dataset. The Raw dataset containes about 13,087 cases. The training and testing numbers are approximately 4.2K and 1K, respectively. This dataset has 32 unique type of event codes. 

\item \textbf{IT incident management Dataset}: This is an anonymized data set extracted from incident management system supporting an enterprise resource planning (ERP) application. It contains 16,000 tickets(process instances) of IT incident management processes. The log contains the life cycle of a ticket. The ticket is opened by a customer. It is acknowledged typically by a team lead, then it gets assigned to a person working on it and after some analysis and other changes, it gets closed. The group that solved the ticket might not correctly resolve the issue. The log contains the name of the last group that solved the ticket. After splitting, the Incident Mgmt. dataset has about 26K training and 6.5K prefix/suffix sequences. This dataset has 32 unique type of event codes. 
\end{itemize}

\subsection{Pre-Processing}


We take each of these datasets and we split the logs into desirable and undesirable instances (by using the performance of each instance against the stated KPIs) and following the langauge modeling approach, only train our models using desirable instances. In the Moodle dataset we filter the dataset, by applying a couple of pre-conditions such that each instance should have at least four distinct states \footnote{https://docs.moodle.org/dev/Process} and no more than 25 state changes.  An undesirable instance examples are chosen with the assumption that bad process instances would shift states back and forth a lot (e.g., issue being reopened multiple times is an Undesirable instance). Hence if more than 25 state changes occur for a given issueID then it would be labelled as an Undesirable instance. Similarly for BPI2012 financial log data we filter cases based on running time. Cases that started in 2012 were filtered out(about 49 percent because they are not likely to finish. Next, we perform performance filtering using total time duration for each case. Cases with a maximum duration of 1 day 19 hours are considered desirable instances while rest of them are labelled as Undesirable performing instances. Each process is a sequence of events and each event is represented by a discrete symbol, which is coded
using the one-hot coding scheme introduced in Section~\ref{subsec:Events/resources-coding-and}.
We randomly divide all processes into 80\% for training and 20\% for
testing. Then, we continue splitting each process in the training and test sets into prefix sequence and suffix sequence such that the minimum prefix length is 4.

\subsection{Experimental Setup and Modeling \label{subsec:Implementation}}

\textit{\emph{For all experiments, deep learning models are implemented
in Tensorflow 1.3.0. Optimizer is Adam \cite{kingma2014adam} with
learning rate of 0.001 and other default parameters. Table~\ref{tab:DCw-MANN-hyper-parameters} describes the hyper-parameter settings, as selected through trial and error.}} To the best of our knowledge, there is no
existing ML based technique for suffix recommendation that considers the problem of outcome based optimal path generation. Therefore, for comparison, we implement custom process-agnostic baselines.  For the datasets (Moodle, Financial Log and Incident Mgmt.), the
baselines are $k$-NN, and GRU. The $k$-NN presents a simple but powerful baseline for the case of vector inputs. Thus it is of interest to see if it works well for sequence inputs as in the case of process analytics. The LSTM, on the hand, has been the state-of-the-art
for this domain, as shown in recent work\cite{evermann2016deep,tax2017predictive}.The GRU is a recent alternative to LSTM, which has been shown to be
equally effective in NLP tasks \cite{cho2014gru}. The $k$-NN works by retrieving $k$ most similar prefixes in the
training data. Then suffix and other desirable outcomes are computed
from the same outcomes of those retrieved cases. The recommendation is either the average of the retrieved outcomes (if continuous), or the
most common outcome (if discrete). 

\textbf{Model Evaluation:} Numerical evaluations with comparisons to baselines play a central role when judging research for most recommender systems, therefore we rely on baseline comparisons/benchmarks to evaluate the quality of recommendations produced by our machinery. Our recommender system provides operational decision support for process users and on a higher level, performs utility optimization. Gunawardana et. al \cite{gunawardana2009survey} provide a survey of evaluation metrics for recommender systems. They observe that the task of optimizing utilities is by far the least explored recommendation task. Hence research/industry standard evaluation metrics do not exist for such task. Prescriptive machine learning models for such tasks are predominantly benchmarked
by matching samples against a reference solution (e.g previously well-performing instances representing ground truth). In our case, we have picked Levenshtein distance metric because it aptly summarizes accuracy/precision in terms of the closeness of recommended sequences to the reference(desired) sequences in high-dimensional vector space. It represents a degree of conformity of evaluated predictions to the true value and is sensitive to differences in error rates, making it effective for judging the effectiveness of our process-aware recommendation machinery. However,
these distances have a quadratic time complexity of the sequence length,
which can be expensive for long sequences. Hence we build a Trie over
the training prefixes for fast retrieval. In our experiments, we choose
$k$ to be $1$ and $5$. We append to the end of each complete process a special token \textless{}END\textgreater{} signaling
its termination. We train the GRU in the same manner as training
a language model \cite{mikolov2012statistical}, which is identical to
next activity prediction. After training, a test prefix will be fed
to the GRU as prior context and the model will continue recommending
the next event step-by-step until the \textless{}END\textgreater{}
symbol is outputted. In our experiment, we use a hidden vector of
size $100$ for both GRU and MANN methods. 


\subsection{Results and Discussion \label{subsec:Next-activity-and}}




For evaluation, we use the edit
distance (Levenshtein distance) as it is a good indication of sequence
similarity where deletion, insertion or substitution are available
as in the case of business processes. To account for variable sequence
lengths, we normalize this distance over the length of the longer
sequence (between 2 sequences). Then, the final metric is calculated as the normalized edit
similarity that equals  $1-$ normalized edit
distance. Consequently, the predicted sequence
is good if its normalized edit similarity to the target sequence is high.

We observe in Table-1 that our MANN based mode outperforms the state of the art LSTM model across all three datasets. The $k$-NN works surprisingly well. However, it
faces some difficulties in this problem of sequence-to-sequence prediction.
First, the prefixes can be slightly different but the suffices can
differ drastically, e.g., due to a single decisive event. Second,
the $k$-NN does not capture the continuation of a process, and thus
suffices from similar instances do not guarantee to be the right continuation.
And third, for $k>1$, there is no easy way to combine multiple suffix
sequences, which shows in the worse result than the case $k=1$.




\begin{table}

\begin{centering}
\begin{tabular}{@{} c | c | c | c @{}}
\hline 
Method & Moodle & fin\_log & Inc\_Mgmt \tabularnewline
\toprule

$5$-NN & 0.817 & 0.588 & 0.418\tabularnewline
$1$-NN & 0.840 & 0.631 & 0.432\tabularnewline
GRU & 0.875 & 0.559 & 0.454\tabularnewline
LSTM & 0.887 & 0.683 & 0.497\tabularnewline
\bottomrule 
MANN & \textbf{0.888} & \textbf{0.691} & \textbf{0.502}\tabularnewline
\hline 
\end{tabular}
\par\end{centering}
\caption{Suffix Recommendation Task: The average normalized edit similarity between
the target suffixes and the suffixes recommended by different models
(higher is better). \label{tab:The-average-normalized}}

\begin{centering}
\begin{tabular}{|l|c|c|c|c|}
\hline 
Hyper-parameters & Moodle & Financial Log & Incident Mgmt. \tabularnewline
\hline 
\# memory slots & 64 & 64 & 64  \tabularnewline
\hline 
Memory slot size & 100 & 64 & 100 \tabularnewline
\hline 
Controller hidden dim & 100 & 100 & 100\tabularnewline
\hline 
\end{tabular}
\par\end{centering}
\caption{MANN hyper-parameters. ({*}) no duplicate.\label{tab:DCw-MANN-hyper-parameters}}

\end{table}

Since the authors in \cite{tax2017predictive} shared, only public the code for the two-layer LSTMs (one is shared-weight), we can only calculate the parameter size for this configuration, which is about 208K trainable parameters. It should be noted that the best model configurations consisting of 3 or 4 layers may have even more than that number of parameters. Our MANN, by contrast, is much simpler with two one-layer controllers and an external memory hence has fewer parameters (less than 125K). This suggests the ability of the external memory to compress and capture essential information in order to perform better. 

Taken together, the results achieved using MANNs demonstrate that our proposed machinery is well-suited for solving both predictive and prescriptive monitoring tasks, with far fewer parameters. We also note that MANNs are relatively new, and we expect that even better performance could be achieved with greater effort in devising encodings for process analytics problems. As well, we have been able to position a range of process analytics problems to leverage future developments/improvements in MANNs. Our approach based on employing labelled datasets should hopefully lead the community to ask a broader range of prescriptive process analytics questions that could be solved using similar machinery as discussed in this paper.

\section{Related Work \label{sec:Related-Work}}

Predictive business process monitoring is a family of techniques concerned with predicting the future state, outcomes and behaviour of ongoing cases of a business process\cite{teinemaa2019outcome}. Relevant to our work, the task of \emph{next activity prediction} and \emph{Process Path Prediction} has been tackled by approaches like state-transition models, hidden Markov models(HMM) and  Probabilistic Finite Automatons(PFA) models \cite{lakshmanan2015markov} \cite{breuker2016comprehensible} \cite{ceci2014completion}. Tax et. al\cite{tax2017predictive} point out that such approaches are \emph{'tailor-made for specific prediction tasks and not readily generalizable'}. Recently, Deep Learning methods such as LSTMs have shown an advantage over such classical methods for making accurate predictions and solving various predictive monitoring tasks \cite{tax2017predictive}. Several survey papers have reviewed the literature on predictive process monitoring. e.g. Marquez-Chamorro et al. \cite{marquez2017predictive} and Di Francescomarino et al. \cite{di2018predictive} classify the literature based on input data, classification algorithm and prediction target.  Similarly, Teinemaa et al. \cite{teinemaa2019outcome} and Verenich et al. \cite{verenich2019survey} also survey the literature by covering various datasets, propose task definitions and provide benchmark comparison of recently proposed algorithms. The output of Predictive business process monitoring techniques, is just \emph{predictions}. Predictions can be used as \emph{early warnings} for taking risk informed decisions but do not explicitly support answering of question like \emph{What action should we take next to achieve a particular goal?} and \emph{Why should we do it?}\cite{LEPENIOTI202057}. Compared to descriptive and predictive business analytics, prescriptive process analytics remains less mature \cite{eili2021systematic}. Marquez et. al. \cite{marquez2017predictive} point out that  \emph{`little attention has been given to providing recommendations'}. Instead of providing specific action recommendations, literature on business process monitoring focuses on forecasting future process events(and outcomes) while leaving the action implementation part to the subjective judgment of process users and business decision makers\cite{dees2019if}. 
Overall, prescriptive business process monitoring techniques \cite{groger2014prescriptive} \cite{conforti2013supporting} \cite{schonenberg2008supporting}  have largely focused on recommending preventive actions in order to support  \emph{risk-informed decision making}.

Eili et al. \cite{eili2021systematic} provide a  systematic review of Recommender Systems in Process Mining and classify recommendation approaches as \emph{`pattern optimization'}, \emph{`risk minimization'}, or \emph{`metric-based'}. They highlight the fact that compared to descriptive and predictive business analytics, prescriptive process analytics remains less mature \cite{eili2021systematic}. Existing prescriptive business process monitoring techniques \cite{groger2014prescriptive} \cite{conforti2013supporting} \cite{schonenberg2008supporting}  are used to recommend  \emph{preventive actions} in order to support  \emph{risk-informed decision making}.  To the best of our knowledge, the problem of recommending best path (representing sequence of activities leading to the process end), based on pre-defined KPIs hasn't been addressed. Closely related to our work, Weinzier et al. \cite{weinzierl2020prescriptive} consider problem of recommending \emph{next best actions} that lead to optimal outcomes. Their work however differs from ours, as their technique relies on explicitly adding control-flow knowledge to their proposed technique via formal process model and uses process simulations to verify and filter the predictions of the trained predictive model. Similarly, Groger et al. \cite{groger2014prescriptive} introduce the concept of recommendation-based business process optimization for data-driven process optimization. Their data-mining driven solution supports adaptive processes and recommends actions for next process step to take for a given process instance in order to avoid performance deviation. Lastly, Schobel et al. \cite{schonenberg2008supporting} propose a technique for early identification of diverging processes that can support operational decision-making processes by for example taking remedial actions as business processes unfold. Overall, prescriptive business process monitoring techniques \cite{groger2014prescriptive} \cite{conforti2013supporting} \cite{schonenberg2008supporting}  are used to recommend  preventive actions in order to support  \emph{risk-informed decision making}. However, compared to above mentioned work which focuses on \emph{early warning recommendations} (e.g predicted metric deviation), our work focuses on \emph{best action recommendations} that maximize the likelihood of achieving desirable outcomes.


\section{Conclusion \label{sec:Conclusions}}

In this paper, we explored the application of recent advances in deep learning for building a \emph{Process-Aware Recommender System}. We investigated a specific type of neural network known as the \emph{memory\textendash augmented neural network (MANN)} for its applications in prescriptive process monitoring tasks. We adapted a recently developed MANN architecture, namely the Differential Neural Computer \cite{graves2016hybrid} and proposed several modifications to the default architecture. We performed evaluations using three labelled datasets to show that our proposed approach performs well 
on the task of suffix recommendation while taking cognisance of the relevant KPIs.
Our future work will involve investigating the behaviour of MANNs on highly complex processes that involve multiple intermediate steps and results, and devising ways to visualise how distant events are remembered and linked together.


\bibliographystyle{plain}




%


\end{document}